\documentclass[letterpaper, 10 pt, conference]{ieeeconf}
\usepackage{cite}
\usepackage{commath}
\usepackage{amsmath}
\usepackage{amssymb}
\usepackage{graphicx}
\usepackage[caption=false, font=footnotesize]{subfig}
\usepackage[misc]{ifsym}
\usepackage{epstopdf}
\usepackage{algorithm}
\usepackage{algorithmic}
\usepackage{siunitx}
\usepackage{dblfloatfix}
\usepackage{tikz}
\usepackage{gensymb}
\usepackage{nicefrac}
\usepackage{xspace}
\usepackage{titlesec}

\IEEEoverridecommandlockouts
\title{A Biologically-Inspired Simultaneous Localization and Mapping\\ System Based on LiDAR Sensor}

\author{
	\normalsize{
	Genghang Zhuang$^1$,
	Zhenshan Bing$^1$,
	Yuhong Huang$^1$,
	Kai Huang$^{2,3}$, and
	Alois Knoll$^1$
	}
	\thanks{\footnotesize{
	$^1$Department\,of\,Informatics, Technical\,University\,of\,Munich, Germany.
	$^2$School of Computer Science and Engineering, Sun\,Yat-sen\,University.
	$^3$Pazhou Laboratory, China.
	}}
}

\begin{document}

\def\LiDAR{Li\textsc{dar}\xspace}
\def\LiDARs{Li\textsc{dar}s\xspace}

\def\figurename{Fig.}
\def\tablename{TABLE}

\setlength{\abovedisplayskip}{5pt}
\setlength{\belowdisplayskip}{5pt}
\titlespacing*{\subsection}{0pt}{0.2\baselineskip}{1pt}

\maketitle

\thispagestyle{empty}
\pagestyle{empty}

\begin{abstract}
Simultaneous localization and mapping (SLAM) is one of the essential techniques and functionalities used by robots to perform autonomous navigation tasks.
Inspired by the rodent hippocampus, this paper presents a biologically inspired SLAM system based on a \LiDAR sensor using a hippocampal model to build a cognitive map and estimate the robot pose in indoor environments.
Based on the biologically inspired models mimicking boundary cells, place cells, and head direction cells, the SLAM system using \LiDAR point cloud data is capable of leveraging the self-motion cues from the \LiDAR odometry and the boundary cues from the \LiDAR boundary cells to build a cognitive map and estimate the robot pose.
Experiment results show that with the \LiDAR boundary cells the proposed SLAM system greatly outperforms the camera-based brain-inspired method in both simulation and indoor environments,
and is competitive with the conventional \LiDAR-based SLAM methods.
\end{abstract}

\IEEEpeerreviewmaketitle

\section{Introduction}

Simultaneous localization and mapping (SLAM), as one of the essential techniques and functionalities used by robots to perform autonomous navigation tasks~\cite{thrun2002probabilistic},
aims to build a map to construct a spatial representation of the unknown environment, and simultaneously locate the robot on the map being built.
A large amount of research focuses on solving the SLAM problem with various types of sensors~\cite{taketomi2017visual,santos2013evaluation,filipenko2018comparison}, including monocular and stereo cameras, RGB-D cameras, and \LiDAR sensors.
Compared to camera sensors, \LiDAR sensors are capable of obtaining more accurate distance and depth information by actively projecting laser beams, and are less subject to illumination changes in the environment. Hence, \LiDAR sensors are widely used in autonomous driving and a considerable amount of research has been performed on using \LiDAR sensors to resolve the SLAM problem~\cite{santos2013evaluation,kohlbrecher2011flexible,hess2016real}, and other autonomous driving tasks including object detection~\cite{yang2018pixor}, classification~\cite{qi2017pointnet}, and tracking~\cite{asvadi20163d}.

In regard to mammals, research findings have shown that animals such as rodents have a different navigation system.
Mammals are born with instinctive abilities and skills to perform navigation and cognition tasks.
Successful discoveries have been made in investigating and understanding the spatial representation and navigation system of mammalian brains.
Studies from neuroscience~\cite{moser2008place} have revealed that the hippocampus and entorhinal cortex play an important role in spatial navigation by coordinating several types of neurons for different functionalities, including place cells~\cite{o1971hippocampus,o1976place}, head direction cells~\cite{taube1990head}, grid cells~\cite{hafting2005microstructure,hasselmo2008grid}, and boundary cells~\cite{hartley2000modeling,lever2009boundary,hinman2019neuronal}, in which the studies demonstrate these types of neurons are related to place recognition, boundary sensing, and path integration which is the capacity to use idiothetic cues to track the animal’s movements~\cite{bing2021toward,byrne2017learning}.

An increasing number of studies have been inspired by the neural mechanisms to solve the navigation~\cite{bing2018end,lechner2020neural,bing2020indirect} and SLAM problem~\cite{milford2004ratslam}. \emph{RatSLAM} in~\cite{milford2004ratslam} utilized place cells and head direction cells to build a hippocampal model to integrate motion data and landmarks from vision sensors when mapping. In addition, a considerable number of related studies have been published based on this navigational model, employing many types of sensors, including cameras~\cite{milford2013brain,zhou2017brain,yu2019neuroslam,milford2008mapping} and RGB-D sensors~\cite{tian2013rgb}. 

\LiDAR sensor, which corresponds to animal echolocation that exists in some bat species and odontocetes, has the intuitive advantage in detecting boundaries in the environment.
In addition, with the advantages of \LiDAR sensors including high accuracy and stability, significant potential exists to further improve the overall performance of the biologically inspired SLAM system by using \LiDAR sensors.
However, to date, little literature has been published with regard to \LiDAR-based biologically inspired SLAM approaches.
Considering the connection to echolocating animals, it is intriguing to explore the applications and investigate the advantages of \LiDAR sensors in biologically inspired navigation.
Due to the developments of neuromorphic chips~\cite{davies2018loihi,furber2014spinnaker}
and the emerging deployments of navigational models~\cite{kreiser2018pose,kreiser2018neuromorphic},
a \LiDAR-based biologically-inspired model with higher performance also has the potential to benefit the navigation system for biomimetic robots.
Hence, in this work we focus on leveraging \LiDAR sensors in biologically-inspired SLAM.

\begin{figure*}[t]
	\vspace{4pt}
    \centering
    \includegraphics[width=0.8\textwidth]{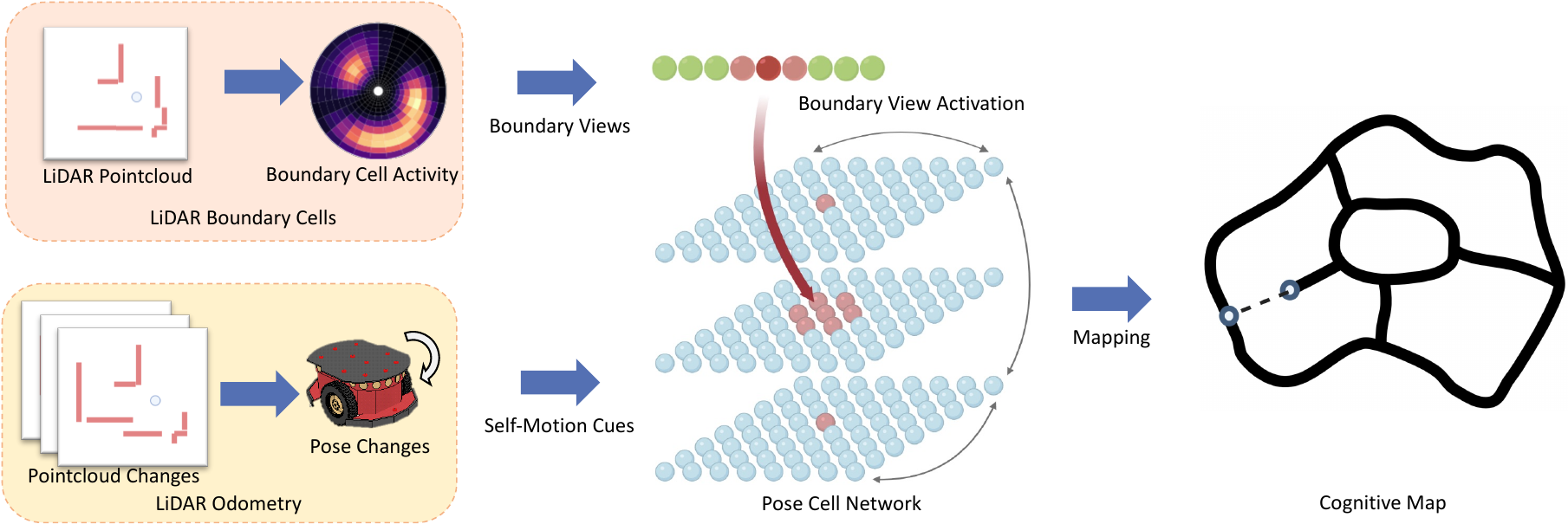}
    \caption{System framework and major modules of the \LiDAR-based biologically inspired SLAM system. The system is composed of the \LiDAR odometry, \LiDAR boundary cells, and the pose cell network.}
    \label{fig:framework}
    \vspace{-12pt}
\end{figure*}

In this paper, we present a biologically inspired SLAM system based on a \LiDAR sensor to build cognitive maps for indoor environments,
which mimics the mechanisms of boundary cells, place cells, and head direction cells of mammals.
 The system, employing a \LiDAR sensor as the primary sensor, consists of three major modules, which include the \LiDAR odometry, \LiDAR boundary cells, and the pose cell network.
The SLAM system using point cloud data from a \LiDAR sensor is capable of leveraging self-motion cues from the \LiDAR odometry and boundary view cues from the \LiDAR boundary cells to build a cognitive map and estimate the robot pose.
Experiment results show that the proposed SLAM system greatly outperforms camera-based RatSLAM in terms of accuracy for SLAM tasks in both simulation and indoor environments,
and is very competitive with the conventional \LiDAR-based SLAM methods.
 Specifically, the main contributions of this work are listed as follows:

\begin{itemize}
	\item To fully leverage the advantages of the \LiDAR sensor, a \LiDAR odometry algorithm is implemented to generate pseudo-odometry data as the self-motion cues for the robot, to reduce the usage of physical odometric sensors such as IMU and wheel encoders that are prone to internal errors and noise.
	\item The egocentric boundary cell model for \LiDAR is implemented to provide boundary view cues to perform place recognition and position calibration. A two-stage boundary view matching approach is proposed to reduce the computational complexity of place recognition.
	\item The pose cell network model is adapted for the \LiDAR sensor to implement the spacial location representation for the robot, and perform path integration and position calibration with the \LiDAR self-motion cues and \LiDAR boundary view cues.
\end{itemize}

\section{Related Work}

\subsection{Conventional LiDAR SLAM}

Error accumulation in point cloud matching and odometry drift is one of the main challenges of \LiDAR-based SLAM.
A number of studies have attempted to tackle this problem.
Kohlbrecher et al.~\cite{kohlbrecher2011flexible} proposed a scan-to-map matching algorithm with multi-resolution maps to reduce the matching errors.
A \LiDAR-based SLAM algorithm proposed by Grisetti et al.~\cite{grisetti2007improved} utilized particle filters to reduce the odometry drift while mapping.
However, 
these algorithms without loop closure detection, which is a technique to recognize previously visited locations, are still subject to cumulative sensor errors over long time mapping.
Hess et al.~\cite{hess2016real} proposed the Cartographer SLAM system, which could perform loop closure detection for \LiDAR SLAM to further improve the accuracy of mapping and localization.
To detect loop closures, the system regularly runs the global pose optimization, in which the \LiDAR scan is matched with all collected submaps to find the closest submap and update the pose estimate. It requires the loop closure detection to complete before the new scan being received.

\subsection{Brain-Inspired Navigation}

A number of studies were investigated to solve the SLAM problem with biologically inspired methods.
Milford et al. proposed \emph{RatSLAM}~\cite{milford2004ratslam} to utilize the place cell and head direction cell model to perform loop closure detection and solve the SLAM problem based on a camera sensor.
Experimental results in~\cite{milford2008mapping,yu2019neuroslam} also show the applicability of their proposed system in large-scale scenarios.
Zhou et al.~\cite{zhou2017brain} employed the ORB features for the camera sensor to improve loop closure detection when mapping.
Tian et al.~\cite{tian2013rgb} proposed an RGB-D sensor-based SLAM method with the path integration model,
and \emph{BatSLAM}~\cite{steckel2013batslam} combined the mapping module of \emph{RatSLAM} and a sonar sensor to solve the SLAM tasks.
However, in the related literature, it is rare that studies have focused on biologically inspired SLAM approaches using \LiDAR sensors.

\section{Methodology}

\newcommand{\pose}{\mathbf{p}}

The proposed biologically inspired SLAM system uses a \LiDAR as the main sensor and data source. The architecture of the SLAM system is illustrated in \figurename\,\ref{fig:framework}. \LiDAR odometry plays an important role in generating motion data for the robot. The boundary cells provides boundary view cues that are processed and integrated based on the \LiDAR observation. The pose cell network estimates the robot 3-DoF pose $\pose = (x, y, \theta)^T$ by performing path integration and loop closure based on the self-motion data from \LiDAR odometry and boundary view cues from the boundary cell module.

\subsection{LiDAR Odometry}
\label{sec:lidar_odom}

\newcommand{\pointcloud}{{P}}
\def\localMap{{\mathbf{M}}}

The \LiDAR odometry in the SLAM system provides odometric motion data for the pose cells to update the network state and perform path integration. Receiving consecutive point clouds from the \LiDAR sensor, the aim of the \LiDAR odometry is to estimate the pose change $\Delta \pose = (\Delta x, \Delta y, \Delta \theta)$ during the time period of the robot movement.

The \LiDAR sensor rotates in the horizontal plane and scans the surrounding objects with a fixed angle increment based on the angular resolution of the \LiDAR sensor. A point cloud $\pointcloud$ from a complete scan can be defined as a sequence of endpoint distances as follows:
\begin{equation}
	P = \big(d_i, i = 1, 2, ..., N \big),
\end{equation}
where $d_i$ is the distance of the $i$-th scan endpoint, and $N$ is the number of points in a complete scan, which is determined by the scanning angular resolution of the \LiDAR sensor.

To estimate the pose change $\Delta \pose$ with a given new input point cloud, the points are first transformed into the form of $\mathbb{R}^2$ Cartesian coordinate $\mathbf{e} = (x,y)^T$ as follows:
\newcommand{\lidarEndpoint}{{\mathbf{e}}}
\newcommand{\scanDis}{{d}}
\newcommand{\scanAngle}{{\alpha}}
\begin{equation}
	\lidarEndpoint_i
  =
  \Bigg(\
  	  \begin{aligned}
  	  \scanDis_i \cdot cos \scanAngle_i&\\
  	  \scanDis_i \cdot sin \scanAngle_i&
  	  \end{aligned}
  	  \
  \Bigg),
\end{equation}
where $\alpha_i$ is the scanning angle of the $i$-th endpoint.

In this work, local \LiDAR mapping on an occupancy grid map is adopted to compute the pose change.
A local occupancy grid map $\localMap$ is employed to downsample the accumulation of \LiDAR observations and reduce the effect of the sensor noise and dynamic obstacles~\cite{elfes1989using}.
The point cloud from \LiDAR observation is matched against the local map to reduce the cumulative drift in scan-to-scan matching~\cite{kohlbrecher2011flexible}.

To estimate the pose change $\Delta\pose$, a nonlinear optimization is constructed and performed
to find the objective pose change $\Delta\pose^\prime$, where the current \LiDAR point cloud can best match the local map $\localMap$ after being projected. The optimization objective is defined as follows:
\begin{equation}
	\Delta \mathbf{p} = \operatorname*{argmax}_{\Delta \mathbf{p}^\prime} \sum_{i}^{N} \localMap^* \Big(\mathbf{R}_{\Delta \mathbf{p}^\prime} \cdot (\mathbf{R}_\mathbf{p} \cdot \lidarEndpoint_i + \mathbf{T}_\mathbf{p}) + \mathbf{T}_{\Delta \mathbf{p}^\prime} \Big),
\end{equation}
where $\pose$ is the current pose of the robot.
$\localMap^*$ is an upsampled local map of $\localMap$ to provide a continuous occupancy function for the discrete grid map $\localMap$~\cite{kohlbrecher2011flexible}.
$\mathbf{R}_\pose$ and $\mathbf{T}_\pose$ are the corresponding rotation and translation transforms for a given pose $\pose = (x, y, \theta)^T$:
\begin{equation}
	\mathbf{R}_\mathbf{p} \cdot \lidarEndpoint + \mathbf{T}_\mathbf{p} = 
	  \begin{pmatrix}
	  cos \theta & -sin \theta\\
	  sin \theta & cos \theta
	  \end{pmatrix}
	  \cdot \lidarEndpoint
	  +
	  \begin{pmatrix}
	  x\\
	  y
	  \end{pmatrix}
	  .
\end{equation}

The input point cloud is initially transformed to the local map with the current pose $\pose$.
The objective $\Delta \mathbf{p}^\prime$ is optimized to yield the pose change estimate $\Delta \pose = (\Delta x, \Delta y, \Delta \theta)$ as the output of \LiDAR odometry, which is subsequently used to update the activity of pose cells and perform path integration.

\subsection{LiDAR Boundary Cells}

\newcommand{\hash}{{h}}
\newcommand{\template}{{B}}
\newcommand{\linear}{{L}}
\newcommand{\view}{{v}}
\newcommand{\viewCells}{{V}}
\newcommand{\similarity}{{s}}

With the \LiDAR observation inputs, the \LiDAR boundary cells module detects and processes the surrounding boundaries into boundary cell activity, which can be considered as abstracted boundary view information for the egocentric scene.
Egocentric boundary views are used to provide the boundary cues to the pose cells to perform loop closure.

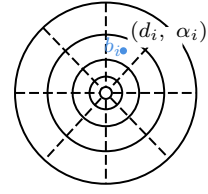
\begin{figure}[t]
	\vspace{4pt}
	\centering
		\tikzset{every picture/.style={line width=0.75pt}} 

\begin{tikzpicture}[x=0.5pt,y=0.5pt,yscale=-1,xscale=1]

\draw [fill={rgb, 255:red, 0; green, 0; blue, 0 }  ,fill opacity=0.76 ] [dash pattern={on 3.75pt off 1.5pt}]  (261.47,362.99) -- (398.65,362.99) ;
\draw  [dash pattern={on 3.75pt off 1.5pt}]  (282.69,313.75) -- (358.88,392.95) -- (377.43,412.24) ;
\draw [color={rgb, 255:red, 0; green, 0; blue, 0 }  ,draw opacity=1 ] [dash pattern={on 3.75pt off 1.5pt}]  (330.06,294.4) -- (330.06,431.58) ;
\draw  [dash pattern={on 3.75pt off 1.5pt}]  (376.03,312.76) -- (331.79,361.11) -- (331.41,361.52) -- (329.91,363.16) -- (284.1,413.23) ;
\draw   (285.93,362.99) .. controls (285.93,338.62) and (305.69,318.86) .. (330.06,318.86) .. controls (354.43,318.86) and (374.19,338.62) .. (374.19,362.99) .. controls (374.19,387.36) and (354.43,407.12) .. (330.06,407.12) .. controls (305.69,407.12) and (285.93,387.36) .. (285.93,362.99) -- cycle ;
\draw   (261.32,363.16) .. controls (261.32,325.28) and (292.03,294.57) .. (329.91,294.57) .. controls (367.79,294.57) and (398.5,325.28) .. (398.5,363.16) .. controls (398.5,401.04) and (367.79,431.75) .. (329.91,431.75) .. controls (292.03,431.75) and (261.32,401.04) .. (261.32,363.16) -- cycle ;
\draw   (304.9,362.99) .. controls (304.9,349.09) and (316.16,337.83) .. (330.06,337.83) .. controls (343.96,337.83) and (355.23,349.09) .. (355.23,362.99) .. controls (355.23,376.89) and (343.96,388.16) .. (330.06,388.16) .. controls (316.16,388.16) and (304.9,376.89) .. (304.9,362.99) -- cycle ;
\draw   (317.7,362.99) .. controls (317.7,356.16) and (323.23,350.62) .. (330.06,350.62) .. controls (336.89,350.62) and (342.43,356.16) .. (342.43,362.99) .. controls (342.43,369.82) and (336.89,375.36) .. (330.06,375.36) .. controls (323.23,375.36) and (317.7,369.82) .. (317.7,362.99) -- cycle ;
\draw  [fill={rgb, 255:red, 255; green, 255; blue, 255 }  ,fill opacity=1 ] (325.55,362.99) .. controls (325.55,360.5) and (327.57,358.48) .. (330.06,358.48) .. controls (332.55,358.48) and (334.57,360.5) .. (334.57,362.99) .. controls (334.57,365.48) and (332.55,367.5) .. (330.06,367.5) .. controls (327.57,367.5) and (325.55,365.48) .. (325.55,362.99) -- cycle ;
\draw  [color={rgb, 255:red, 255; green, 255; blue, 255 }  ,draw opacity=1 ][fill={rgb, 255:red, 255; green, 255; blue, 255 }  ,fill opacity=1 ] (347.21,313.81) -- (387.25,313.81) -- (387.25,327.87) -- (347.21,327.87) -- cycle ;
\draw  [color={rgb, 255:red, 74; green, 144; blue, 226 }  ,draw opacity=1 ][fill={rgb, 255:red, 74; green, 144; blue, 226 }  ,fill opacity=1 ] (341.34,331.1) .. controls (341.34,330.02) and (342.22,329.14) .. (343.31,329.14) .. controls (344.39,329.14) and (345.27,330.02) .. (345.27,331.1) .. controls (345.27,332.19) and (344.39,333.07) .. (343.31,333.07) .. controls (342.22,333.07) and (341.34,332.19) .. (341.34,331.1) -- cycle ;

\draw (346.18,326.67) node [anchor=south west] [inner sep=0.75pt]  [font=\footnotesize]  {$( d_{i} ,\ \alpha _{i})$};
\draw (335.57,327.97) node  [font=\footnotesize,color={rgb, 255:red, 74; green, 144; blue, 226 }  ,opacity=1 ]  {$b_{i}$};
\end{tikzpicture}
		\vspace{-4pt}
	\caption{Receptive fields of boundary cells.}
	\label{fig:bc_scheme}
	\vspace{-16pt}
\end{figure}

The computational model of boundary cells was first predicted in~\cite{hartley2000modeling,barry2006boundary}, and the existence of boundary cells was discovered and proposed in~\cite{lever2009boundary,hinman2019neuronal}.
\newcommand{\bcActivity}{{b}}
Each neuron of the boundary cells responds to a boundary in proximity in a certain area, which is denoted as the receptive field.
In the \LiDAR boundary cell model, the receptive fields of the cells are arranged around the robot in concentric rings at different distances, as illustrated in \figurename{\,\ref{fig:bc_scheme}}.
Each boundary cell is tuned to fire maximally with the input at the center of its receptive field.
Given a \LiDAR point input $\lidarEndpoint = (\scanDis, \scanAngle)$ in the polar coordinate system,
the neuron activity update $\Delta\bcActivity_i$ for the $i$-th boundary cell, whose receptive field center is at $(\scanDis_i, \scanAngle_i)$, is defined as follow:
\begin{equation}
	\Delta\bcActivity_i = \frac{1}{\scanDis} \cdot e^{-(\frac{\scanDis-\scanDis_i}{\sigma_{\scanDis_i}})^2} \cdot e^{-(\frac{\scanAngle-\scanAngle_i}{\sigma_\scanAngle})^2},
\end{equation}
where $\sigma_{\scanDis_i}$ and $\sigma_\scanAngle$ are the parameters to determine the area sensitivity of the receptive field by tuning the variances for angle and distance. $\sigma_{\scanDis_i}$ is a variable linear to $\scanDis_i$ in order to expand the receptive field when distance increases~\cite{lever2009boundary},
while $\sigma_\scanAngle$ is a constant determined by the number of boundary cells in a ring.
The activity level of boundary cells is inversely proportional to the distance to corresponds the property that the boundary cells' firing rate gradually increases with approaching closer to the boundary~\cite{lever2009boundary}.

The real-time \LiDAR boundary cell activity will be compared with the previously learned boundary views to generate a pose calibration activity if a matched previous view is found, or to learn a new boundary view if none is matched.
\newcommand{\bcSize}{{M}}
To reduce the matching complexity, a two-stage boundary view matching approach is proposed in this work.
In the first stage, coarse prematching is performed to boost the real-time performance and reduce the computational complexity.
A coarse view feature for the \LiDAR boundary cells is proposed to roughly describe and index the boundary views by utilizing the aggregate of the boundary cell firing activity. The coarse view feature $\hash$ is defined as follows:
\begin{equation}
	\hash = \Bigl\lfloor 10^{-d_s} \cdot \sum\limits^{\bcSize}_{i}\limits{\bcActivity_i} \Bigr\rfloor,
\end{equation}
where $d_s$ is a constant downscaling factor to control the preciseness for the coarse matching, and $\bcSize$ is the size of the boundary cells.
The coarse view feature $\hash$ can be considered a hash value for the \LiDAR boundary view, which is leveraged to reduce unnecessary matching processes in the second stage.
By combining the coarse feature, a \LiDAR boundary view $\view$ is constructed as a tuple of the coarse view feature $\hash$ and the boundary cell activity $\template$:
\begin{equation}
	\view = \big< \hash, \template=\{\bcActivity_i,i=1,2,...,\bcSize\} \big>.
\end{equation}

\newcommand{\similarityFunc}{S}
In the second stage of the boundary view matching process, 
for the boundary views with the equal coarse view feature $\hash$, the corresponding views are subsequently compared to estimate the similarity $\similarity=\similarityFunc(\template_i, \template)$, which is evaluated by computing the mean squared differences between the two boundary views:
\begin{equation}
	\similarityFunc(\template_1,\template_2) = \frac{1}{\bcSize} \sum_{i}^{\bcSize}{\big(\template_{1}(i) - \template_{2}(i) \big)^2},
\end{equation}
The similarity $\similarity$ is then used to determine whether the learned boundary view $\view_i$ is matched with the current boundary view $\view$.
The boundary views are subsequently associated with the boundary view vector $\viewCells$, which is defined as a vector of matching activity levels of the boundary views.
To maintain the matching levels over time for each boundary view, the activity of each node in the vector will be updated as follows:

\begin{equation}
	\viewCells_i = 1 - \frac{min\big(s_t, S(\template_i, \template)\big)}{s_t},
\end{equation}
where $s_t$ is the matching threshold for similarity.
The activity level is calculated based on the matching error between the boundary view $\view=\big<\hash,\template\big>$ for the current \LiDAR observation and the $i$-th learned boundary view $\view_i=\big<\hash_i,\template_i\big>$, in which the error is clamped, inverted, and scaled to $[0,1]$. In the event that no present boundary view is matched, the vector $\viewCells$ will be extended to $\viewCells_{i+1}$ to associate the new boundary view.
Eventually, the activity $\viewCells$ is output to the pose cell network to perform pose association for a new boundary view or loop closure for learned boundary views.

\subsection{Pose Cell Network}

\newcommand{\pc}{{\mathbf{PC}}}
\newcommand{\expNode}{{e}}
\newcommand{\link}{{l}}
\newcommand{\conMatrix}{{\mathbf{A}}}

In this work, we adapt and implement the 3D pose cell network that combines 2D place cells and 1D head direction cells to maintain the pose representation and integrate self-motion cues from \LiDAR odometry and the boundary view cues from \LiDAR boundary cells, which is designed to reduce odometry drift and solve boundary view ambiguity in the process of mapping. 
Leveraging the pose cell network enables the proposed SLAM system to build a cognitive map by performing path integration based on self-motion cues.
In addition, with the \LiDAR view cues, the pose cell network can calibrate the estimated pose and the online cognitive map by performing loop closure to reduce the accumulated errors and drifts by \LiDAR odometry.

The pose cell network is a 3D continuous attractor network (3D-CAN)~\cite{samsonovich1997path}, which can be represented as a 3D matrix of the activity: $\pc_{x^\prime ,y^\prime, \theta^\prime}$.
The three dimensions of the pose cell network represent the three degrees of the 3-DoF pose $\pose = (x, y, \theta)^T$, respectively, where $x^\prime ,y^\prime, \theta^\prime \in \mathbb{Z}$ are the discrete representation of $x,y,\theta \in \mathbb{R}$.
Each pose cell unit in the pose cell network is connected with its neighbor units with excitatory and inhibitory connections, which wrap across the boundaries of the network in three dimensions to enable the pose cell network to represent an unbounded space with a limited number of pose cells~\cite{milford2004ratslam}.
The pose cell network enrolls local excitation and global inhibition activities, which are based on a three-dimensional Gaussian distribution, to self-update the pose cell network dynamics over time~\cite{milford2008mapping}. The stable state of the pose cell network, in which the activated cells are clustered, encodes the estimate of pose $\pose$ as the centroid of the activity packet~\cite{ball2013openratslam}.

Driven by the robot motion, the path integration updates the activity of the pose cells based on the self-motion cues from the \LiDAR odometry detailed in \textsc{Section}~\ref{sec:lidar_odom}.
The activity of each pose cell is shifted along with the translational and rotational movement based on the pose change $\Delta \pose = (\Delta x, \Delta y, \Delta \theta)$.
The activity update for each pose cell is defined as follows:
\begin{equation}
\begin{gathered}
	\Delta \pc_{x^\prime ,y^\prime, \theta^\prime} = 
	\sum^{\delta_{x}+1}_{i=\delta_{x}}
	\sum^{\delta_{y}+1}_{j=\delta_{y}}
	\sum^{\delta_{\theta}+1}_{k=\delta_{\theta}} { r_{ijk} \cdot \pc_{(x^\prime+i),(y^\prime+j),(\theta^\prime+k)} } \\
	\delta_{x} = \lfloor k_x\Delta x \rfloor,
	\delta_{y} = \lfloor k_y\Delta y \rfloor,
	\delta_{\theta} = \lfloor k_\theta\Delta \theta \rfloor,
\end{gathered}
\end{equation}
where $r_{ijk}$ is a residue based on the fractional part of the pose changes spread over the $2\times 2\times 2$ cube to reduce the precision loss by quantization~\cite{milford2004ratslam}, $k_x, k_y, k_\theta$ are the constant scaling factors for the three dimensions.

Given the \LiDAR boundary view cues from the boundary cells, a calibrating activity is injected into the pose cell network to perform further loop closure and re-localization.
The activity of each pose cell is updated based on the activity of boundary cells $\viewCells$, as defined as follows:
\begin{equation}
	\Delta \pc_{x^\prime ,y^\prime, \theta^\prime} = 
	k_\viewCells \cdot \sum_{i}{\conMatrix_{i, x^\prime ,y^\prime, \theta^\prime} \cdot \viewCells_i},
\end{equation}
where $k_\viewCells$ is the constant calibration rate, and $\conMatrix_{i, x^\prime ,y^\prime, \theta^\prime}$ is an adjacency matrix for the connections from the boundary cells to the pose cells. When a new boundary view $\viewCells_{i}$ is learned, an excitatory link from $\viewCells_{i}$ to the current state of the pose cell network $(x^\prime ,y^\prime, \theta^\prime)$ is established, where $\conMatrix_{i, x^\prime ,y^\prime, \theta^\prime}$ is accordingly set to $1$.
To solve the boundary view ambiguity, only a sequence of updates by consecutive boundary views can shift the main activity packet of the pose cells.

In the mapping process, the information from the \LiDAR odometry, \LiDAR boundary cells, and the pose cells are combined and accumulated to estimate the robot pose $\pose = (x, y, \theta)^T$ in the $\mathbb{R}^3$ space, and build a cognitive map as a topological graph of robot movement experiences.
A node in the cognitive map is an experience node defined as a tuple of the states of the pose cells $\pc$, the boundary cells $\viewCells$, and the pose estimate $\pose$:
\begin{equation}
	\expNode_{i} = \big< \pc^i, \viewCells^i, \pose^i \big>.
\end{equation}
After a period of robot movement, when a new boundary view is learned, a new experience node $\expNode^j$, separate from the previous node $\expNode^i$, is created based on the pose change $\Delta \pose^{ij}$.
Then the two experience nodes are connected with a directed edge from $\expNode^i$ to $\expNode^j$ based on the pose transition $\link_{ij}$, which is defined as follows:
\begin{equation}
\begin{gathered}
	\expNode_{j} = \big< \pc^j, \viewCells^j, \pose^i + \Delta \pose^{ij} \big>,\\
	\link_{ij}= \big< \Delta \pose^{ij}, \Delta t^{ij} \big>.
\end{gathered}
\end{equation}
When a learned boundary view is observed and loop closure is detected, a new transition between two existing experience nodes is established, which results in a closure in the cognitive map. The pose transition $\Delta \pose$ of each node and edge is accordingly updated as follows to distribute the accumulated odometry error over the loop trail, and update the pose estimate $\pose = (x, y, \theta)^T$ of the robot.

\small
\begin{equation}
	\Delta \pose^i = a\Bigg[ \sum_{j=1}^{N_f}{\Big(\pose^j-\pose^i-\Delta \pose^{ij}\Big)} + \sum_{k=1}^{N_t}{\Big(\pose^k-\pose^i-\Delta \pose^{ki}\Big)} \Bigg],
\end{equation}
\normalsize
where $a$ is a constant correction factor set to $0.5$, $N_f$ and $N_t$ are the number of the outgoing and incoming edges of $\expNode_i$.
Hence, the cognitive map and the pose estimate $\pose$ are obtained and updated online as the output of the proposed \LiDAR-based biologically inspired SLAM system.

\begin{figure*}[t]
	\centering
	\subfloat[Original scenario for camera]{
		\includegraphics[width=0.2\linewidth]{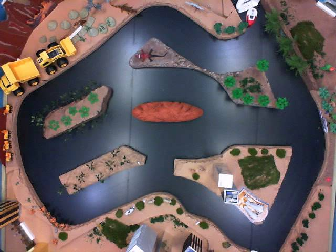}
	}\hfil
	\subfloat[Rebuilt maze scenario for \LiDAR simulation]{
		\includegraphics[width=0.2\linewidth]{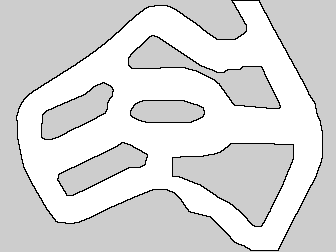}
	}\hfil
	\subfloat[Rebuilt maze in 3D view]{
		\includegraphics[width=0.2\linewidth]{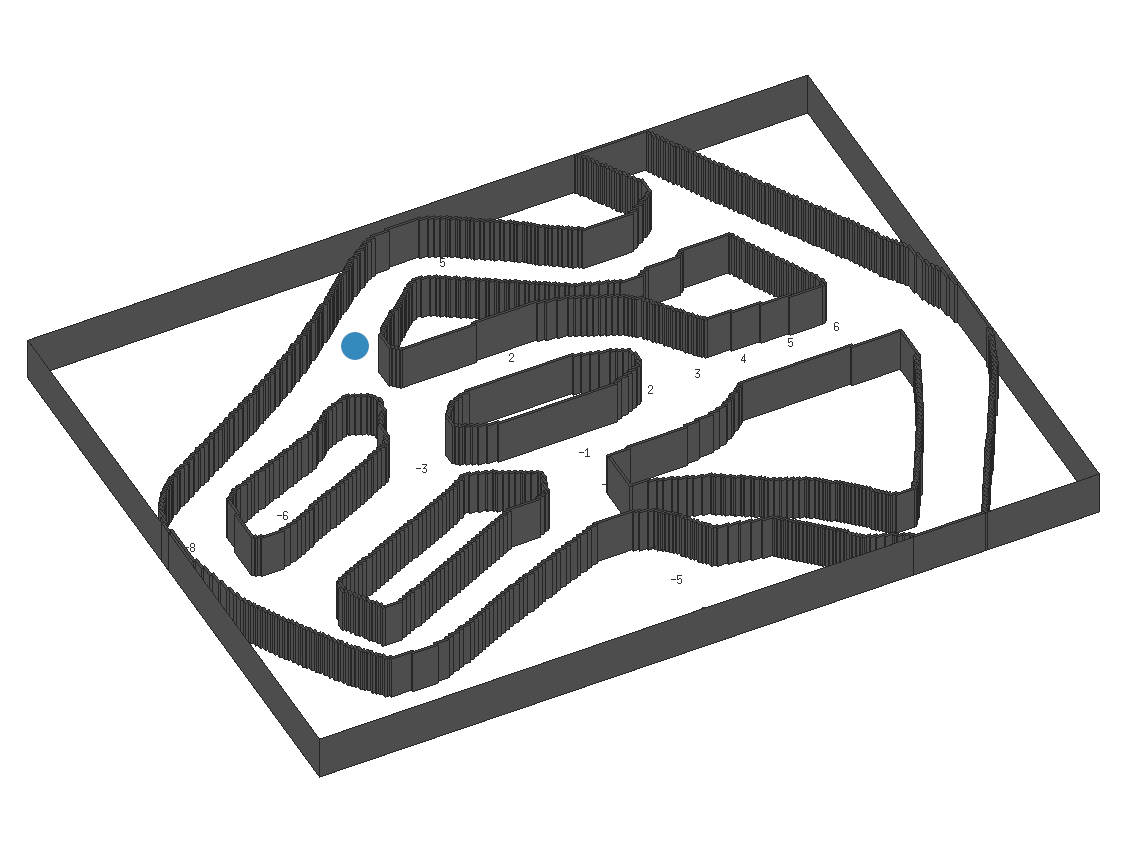}
	}\hfil
	\subfloat[\LiDAR point cloud sample]{
		\includegraphics[width=0.2\linewidth]{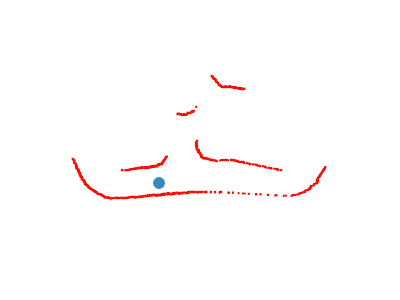}
	}
	
	\caption{Overview of the maze simulation scenario.}
	\label{fig:irat-scenario-overview}
	\vspace{-10pt}
\end{figure*}

\begin{figure*}[t]
	\centering
	\subfloat[t~=~\SI{200}{s}]{
		\includegraphics[width=0.2\linewidth]{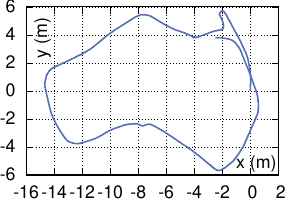}
	}\hfil
	\subfloat[t~=~\SI{400}{s}]{
		\includegraphics[width=0.2\linewidth]{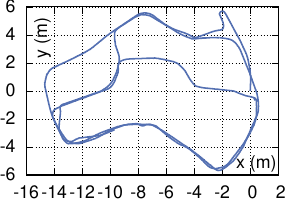}
	}\hfil
	\subfloat[t~=~\SI{600}{s}]{
		\includegraphics[width=0.2\linewidth]{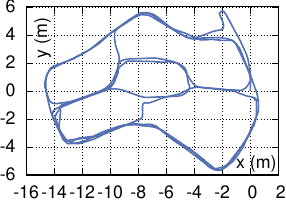}
	}\hfil
	\subfloat[Ground truth]{
		\includegraphics[width=0.2\linewidth]{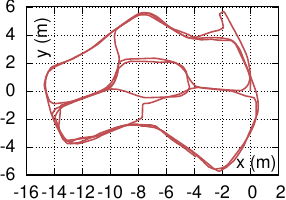}
	}
	
	\caption{Mapping progress in the maze simulation scenario.}
	\label{fig:irat-scenario-mapping-progress}
	\vspace{-10pt}
\end{figure*}

\section{Experiments}

In this section, we present the experimental results for the proposed approach.
To evaluate the \LiDAR-based biologically inspired SLAM system, experiments are conducted in both simulation and real indoor scenarios to investigate the mapping and localization performance.
The proposed SLAM system is implemented and tested on the \emph{Robot Operating System} (ROS)~\cite{quigley2009ros}.
We first report the experimental results of a simulation experiment to evaluate the mapping performance as well as the localization accuracy.
Second, a mapping experiment in an indoor environment based on a ground robot is detailed. The results of the indoor experiment are illustrated and analyzed to demonstrate the applicability and accuracy of the proposed system in the real world.

\subsection{Simulation Experiment}

In the simulation experiments, a maze scenario is ported from the \emph{iRat 2011 Australia} dataset~\cite{ball2013openratslam}, which is originally designed for the visual \emph{RatSLAM}. Since the original dataset does not include \LiDAR sensor data, it is not feasible to carry out the \LiDAR SLAM experiments using the dataset. Therefore, we rebuild the maze scenario in the simulator to make it suited for \LiDAR-based simulations.

\figurename~\ref{fig:irat-scenario-overview} gives an overview of the maze scenario.
Utilizing the \emph{Stage} simulator in ROS, the point cloud data for a robot spawned in the maze scenario are obtained from a configurable virtual \LiDAR in the simulator. At the same time, the ground truth information of the poses and movements of robots are also collected for evaluation purposes, but the odometry data is not involved in the SLAM process. Only the \LiDAR point cloud data are used in the proposed method.

In the simulation experiment, the robot is controlled manually to navigate the maze for 600 seconds to evaluate the mapping performance with loop closure during long-term mapping.
\figurename~\ref{fig:irat-scenario-mapping-progress} illustrates the progressive cognitive maps built by the SLAM system in the simulation experiment and the ground truth path provided by the ROS Stage simulator.
As shown in the mapping progress over time, the error in the cognitive map remains low over time.
The mapping result of high similarity to the ground truth path demonstrates that SLAM performance is sufficiently accurate.

\begin{figure}[t]
	\centering
		\includegraphics[width=0.95\linewidth]{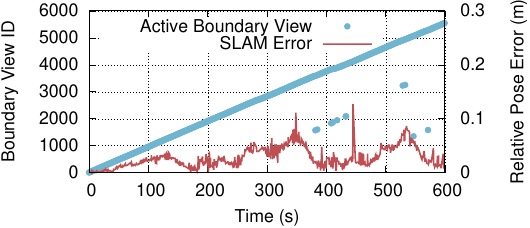}
		\vspace{-6pt}
	\caption{Active boundary cell and relative pose error over time in simulation.}
	\label{fig:irat-local-view}
	\vspace{-16pt}
\end{figure}

\begin{figure*}[t]
	\centering
	\subfloat[Indoor maze scenario]{
    \includegraphics[width=0.18\linewidth]{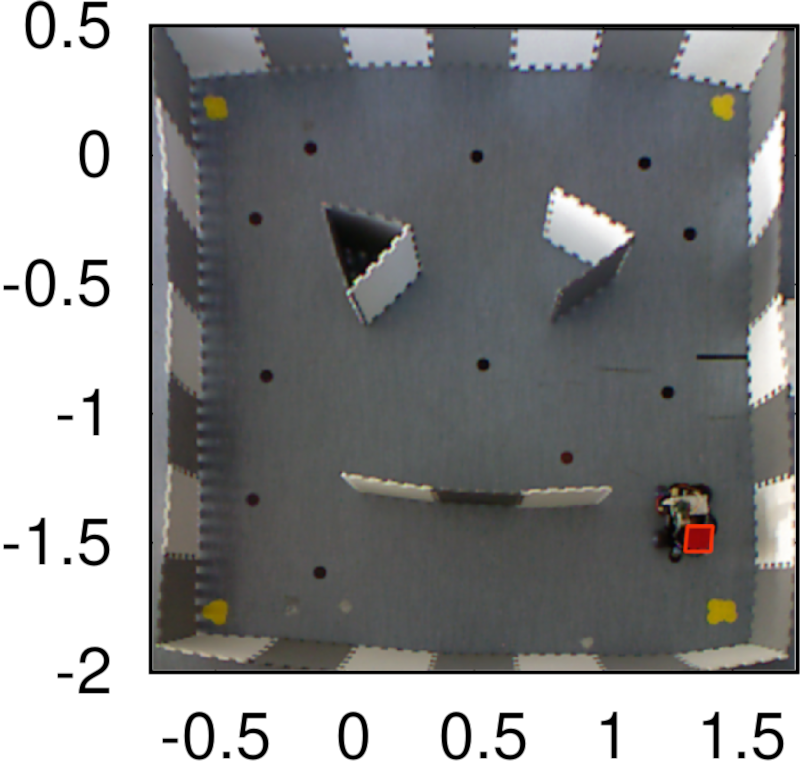}
    \label{fig:maze-bird-view}
    }\hfil
	\subfloat[t~=~\SI{200}{s}]{
		\includegraphics[width=0.18\linewidth]{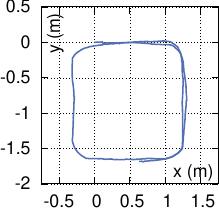}
		\label{fig:garching-hb-mapping-progress-1}
	}\hfil
	\subfloat[t~=~\SI{400}{s}]{
		\includegraphics[width=0.18\linewidth]{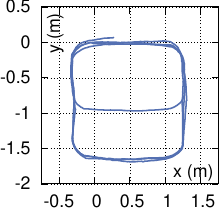}
		\label{fig:garching-hb-mapping-progress-2}
	}\hfil
	\subfloat[t~=~\SI{800}{s}]{
		\includegraphics[width=0.18\linewidth]{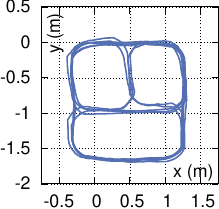}
		\label{fig:garching-hb-mapping-progress-3}
	}\hfil
	\subfloat[Ground truth]{
		\includegraphics[width=0.18\linewidth]{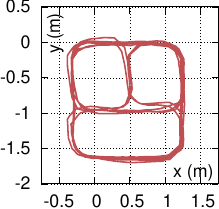}
	}
	\caption{Mapping progress in the real indoor maze environment.}
	\label{fig:garching-hb-mapping-progress}
	\vspace{-10pt}
\end{figure*}

\figurename~\ref{fig:irat-local-view} reveals the activity of the boundary cells, which further reflects the loop closure detection and calibration of the cognitive map and the robot pose estimate.
The number of boundary cells increases as the robot keeps moving to unexplored scenes, which are learned as the new boundary views.
This is consistent with \figurename~\ref{fig:irat-local-view}, in which the increase of the boundary view ID composes the main part of the scatter plot.
The figure also reveals the loop closure detection during the mapping progress.
The dramatic decrease in the active boundary view ID indicates that a number of previously learned boundary views are matched with the current \LiDAR observation.
Combining \figurename~\ref{fig:irat-scenario-mapping-progress} and \figurename~\ref{fig:irat-local-view}, it is obvious that during the mapping process, at around \SI{380}{s} the proposed SLAM managed to detect the previously learned views and performed loop closure calibration to reduce the accumulated error in the map,
which demonstrates the ability of the boundary cells to detect loop closure in the mapping process.

In \figurename~\ref{fig:irat-local-view} we also report the relative pose error (RPE)~\cite{kummerle2009measuring} during the mapping process in the simulation experiment. The localization RPE computed based on the ground truth data is up to \SI{0.127}{m} at around \SI{440}{s}, with a mean value of \SI{0.0284}{m} for the 600 seconds of the mapping process in the $\SI{16.8}{m} \times \SI{12.6}{m}$ space. As shown in the graph, the error remains at a low level for the whole mapping process. For several periods of time around \SI{250}{s} to \SI{350}{s}, and \SI{480}{s} to \SI{540}{s}, the position error started to accumulate but was soon corrected and remained low for a period of time. The transition of the accumulated error shows the process of loop closure detection and demonstrates the high performance of the proposed SLAM system for mapping and localization.

\subsection{Indoor Experiment}

To further evaluate the performance of the proposed SLAM system, experiments were conducted in a real indoor scenario, based on a ground robot platform.
The ground robot is driven by four differential wheels which are controlled by an embedded \emph{Raspberry Pi 3} single-board computer. A 2D \emph{Hokuyo UTM-30LX-EW} \LiDAR sensor is installed on the robot to obtain \LiDAR point cloud for mapping experiments.

To specifically evaluate the mapping accuracy in the experiments,
as shown in \figurename~\ref{fig:maze-bird-view},
we built a maze experimental environment with an external localization and tracking system based on a bird view camera. Due to the limitations of the field of view and installation height for the camera, the size of the maze is relatively limited.
In the indoor maze experiment, the robot was controlled to continuously navigate the maze for a specified period of time,
while the bird view camera kept tracking the movements of the robot in real time to provide the ground truth for localization evaluation in parallel.

\begin{figure}[t]
	\centering
		\includegraphics[width=0.95\linewidth]{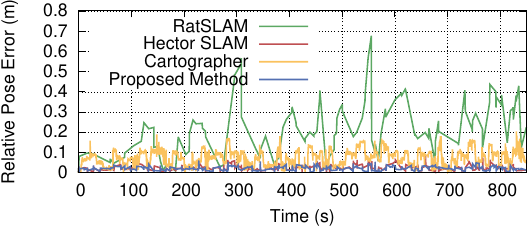}
		\vspace{-4pt}
	\caption{Relative pose errors over time in the maze mapping process.}
	\label{fig:garching-hb-errors}
	\vspace{-16pt}
\end{figure}

The progress of the mapping experiment for \SI{900}{s} in the maze settings is illustrated in \figurename~\ref{fig:garching-hb-mapping-progress} from \figurename~\ref{fig:garching-hb-mapping-progress-1} to \figurename~\ref{fig:garching-hb-mapping-progress-3}, in which each cognitive map was taken at the annotated intervals during the mapping process.
To further quantify the localization and mapping performance, in the indoor experiments, we additionally introduce two related SLAM algorithms for comparison purposes.
To date, since it is rare that studies have focused on \LiDAR-based biologically inspired SLAM algorithms, we evaluated \emph{RatSLAM}~\cite{ball2013openratslam}, which is based on a camera sensor and is one of the most commonly used open-source biologically-inspired SLAM system.
In addition, the experiments also include two conventional \LiDAR-based SLAM systems named \emph{Hector SLAM}~\cite{kohlbrecher2011flexible} and Cartographer~\cite{hess2016real}, which are ones of the state-of-art indoor SLAM methods for indoor scenarios~\cite{santos2013evaluation,filipenko2018comparison}.
We mainly evaluated the SLAM accuracy for the involved SLAM methods in the same experimental settings. In the experiments, relative pose errors (RPE) are computed based on the estimated poses given by the SLAM algorithms and the real-time ground-truth poses given by the bird view camera.
\figurename~\ref{fig:garching-hb-errors} shows the position RPE curves of the proposed method, \emph{RatSLAM}, \emph{Hector SLAM}, and \emph{Cartographer}, during the mapping process of the indoor maze environment.
As shown in the figure, the error curve of \emph{RatSLAM} fluctuated more dramatically than that of the proposed biologically-inspired method and the others.
The position error of the proposed SLAM system is much stabler and lower than \emph{RatSLAM} and is close to the state-of-art conventional SLAM methods.
Interestingly, in this particular small-scale scenario, the error of \emph{Cartographer} is slightly higher than \emph{Hector SLAM}.
The Root Mean Square Error (RMSE) for RPE of the proposed method for the estimated poses in the indoor experiments is \SI{2.23}{cm}. In comparison, the RMSE of \emph{RatSLAM}, \emph{Hector SLAM}, \emph{Cartographer} are \SI{25.38}{cm}, \SI{2.59}{cm}, and \SI{7.97}{cm}.
The results demonstrate that the accuracy of the proposed \LiDAR-based biologically-inspired SLAM method outperforms camera-based \emph{RatSLAM} and the performance is close to and slightly higher than that of the conventional \LiDAR SLAM method \emph{Hector SLAM}.

\section{Conclusion}

In this paper, we present a novel biologically-inspired SLAM system that mimics the mechanisms of boundary cells, place cells, and head direction cells, and leverages the \LiDAR sensor to perform cognitive mapping and localization for indoor environments.
With the self-motion cues and the boundary view cues from the \LiDAR odometry and \LiDAR boundary cells, the proposed SLAM system performs path integration and loop closure in the pose cell network to maintain and calibrate the robot pose estimates and build a cognitive map.
Experimental results in simulation and the realistic indoor environment show that the proposed SLAM system is highly applicable and accurate in both scenarios and greatly outperforms RatSLAM,
and is competitive with the state-of-art conventional \LiDAR-based SLAM methods.

\section*{Acknowledgments}

\normalsize{
This research has received funding from the Pazhou Laboratory Grants with \texttt{No.\,PZL2021KF0020}.
}

\IEEEtriggeratref{21}
\bibliographystyle{IEEEtran}
\bibliography{ref}

\begin{thebibliography}{10}
\providecommand{\url}[1]{#1}
\csname url@samestyle\endcsname
\providecommand{\newblock}{\relax}
\providecommand{\bibinfo}[2]{#2}
\providecommand{\BIBentrySTDinterwordspacing}{\spaceskip=0pt\relax}
\providecommand{\BIBentryALTinterwordstretchfactor}{4}
\providecommand{\BIBentryALTinterwordspacing}{\spaceskip=\fontdimen2\font plus
\BIBentryALTinterwordstretchfactor\fontdimen3\font minus
  \fontdimen4\font\relax}
\providecommand{\BIBforeignlanguage}[2]{{%
\expandafter\ifx\csname l@#1\endcsname\relax
\typeout{** WARNING: IEEEtran.bst: No hyphenation pattern has been}%
\typeout{** loaded for the language `#1'. Using the pattern for}%
\typeout{** the default language instead.}%
\else
\language=\csname l@#1\endcsname
\fi
#2}}
\providecommand{\BIBdecl}{\relax}
\BIBdecl

\bibitem{thrun2002probabilistic}
S.~Thrun, ``Probabilistic robotics,'' \emph{Communications of the ACM},
  vol.~45, no.~3, pp. 52--57, 2002.

\bibitem{taketomi2017visual}
T.~Taketomi, H.~Uchiyama, and S.~Ikeda, ``Visual slam algorithms: a survey from
  2010 to 2016,'' \emph{IPSJ Transactions on Computer Vision and Applications},
  vol.~9, no.~1, p.~16, 2017.

\bibitem{santos2013evaluation}
J.~M. Santos, D.~Portugal, and R.~P. Rocha, ``An evaluation of 2d slam
  techniques available in robot operating system,'' in \emph{2013 IEEE
  International Symposium on Safety, Security, and Rescue Robotics
  (SSRR)}.\hskip 1em plus 0.5em minus 0.4em\relax IEEE, 2013, pp. 1--6.

\bibitem{filipenko2018comparison}
M.~Filipenko and I.~Afanasyev, ``Comparison of various slam systems for mobile
  robot in an indoor environment,'' in \emph{2018 International Conference on
  Intelligent Systems (IS)}.\hskip 1em plus 0.5em minus 0.4em\relax IEEE, 2018,
  pp. 400--407.

\bibitem{kohlbrecher2011flexible}
S.~Kohlbrecher, O.~Von~Stryk, J.~Meyer, and U.~Klingauf, ``A flexible and
  scalable slam system with full 3d motion estimation,'' in \emph{2011 IEEE
  international symposium on safety, security, and rescue robotics}.\hskip 1em
  plus 0.5em minus 0.4em\relax IEEE, 2011, pp. 155--160.

\bibitem{hess2016real}
W.~Hess, D.~Kohler, H.~Rapp, and D.~Andor, ``Real-time loop closure in 2d lidar
  slam,'' in \emph{2016 IEEE International Conference on Robotics and
  Automation (ICRA)}.\hskip 1em plus 0.5em minus 0.4em\relax IEEE, 2016, pp.
  1271--1278.

\bibitem{yang2018pixor}
B.~Yang, W.~Luo, and R.~Urtasun, ``Pixor: Real-time 3d object detection from
  point clouds,'' in \emph{Proceedings of the IEEE conference on Computer
  Vision and Pattern Recognition}, 2018, pp. 7652--7660.

\bibitem{qi2017pointnet}
C.~R. Qi, H.~Su, K.~Mo, and L.~J. Guibas, ``Pointnet: Deep learning on point
  sets for 3d classification and segmentation,'' in \emph{Proceedings of the
  IEEE conference on computer vision and pattern recognition}, 2017, pp.
  652--660.

\bibitem{asvadi20163d}
A.~Asvadi, P.~Gir{\~a}o, P.~Peixoto, and U.~Nunes, ``3d object tracking using
  rgb and lidar data,'' in \emph{2016 IEEE 19th International Conference on
  Intelligent Transportation Systems (ITSC)}.\hskip 1em plus 0.5em minus
  0.4em\relax IEEE, 2016, pp. 1255--1260.

\bibitem{moser2008place}
E.~I. Moser, E.~Kropff, and M.-B. Moser, ``Place cells, grid cells, and the
  brain's spatial representation system,'' \emph{Annu. Rev. Neurosci.},
  vol.~31, pp. 69--89, 2008.

\bibitem{o1971hippocampus}
J.~O'Keefe and J.~Dostrovsky, ``The hippocampus as a spatial map: Preliminary
  evidence from unit activity in the freely-moving rat.'' \emph{Brain
  research}, 1971.

\bibitem{o1976place}
J.~O'Keefe, ``Place units in the hippocampus of the freely moving rat,''
  \emph{Experimental neurology}, vol.~51, no.~1, pp. 78--109, 1976.

\bibitem{taube1990head}
J.~S. Taube, R.~U. Muller, and J.~B. Ranck, ``Head-direction cells recorded
  from the postsubiculum in freely moving rats. i. description and quantitative
  analysis,'' \emph{Journal of Neuroscience}, vol.~10, no.~2, pp. 420--435,
  1990.

\bibitem{hafting2005microstructure}
T.~Hafting, M.~Fyhn, S.~Molden, M.-B. Moser, and E.~I. Moser, ``Microstructure
  of a spatial map in the entorhinal cortex,'' \emph{Nature}, vol. 436, no.
  7052, pp. 801--806, 2005.

\bibitem{hasselmo2008grid}
M.~E. Hasselmo, ``Grid cell mechanisms and function: contributions of
  entorhinal persistent spiking and phase resetting,'' \emph{Hippocampus},
  vol.~18, no.~12, pp. 1213--1229, 2008.

\bibitem{hartley2000modeling}
T.~Hartley, N.~Burgess, C.~Lever, F.~Cacucci, and J.~O'keefe, ``Modeling place
  fields in terms of the cortical inputs to the hippocampus,''
  \emph{Hippocampus}, vol.~10, no.~4, pp. 369--379, 2000.

\bibitem{lever2009boundary}
C.~Lever, S.~Burton, A.~Jeewajee, J.~O'Keefe, and N.~Burgess, ``Boundary vector
  cells in the subiculum of the hippocampal formation,'' \emph{Journal of
  Neuroscience}, vol.~29, no.~31, pp. 9771--9777, 2009.

\bibitem{hinman2019neuronal}
J.~R. Hinman, G.~W. Chapman, and M.~E. Hasselmo, ``Neuronal representation of
  environmental boundaries in egocentric coordinates,'' \emph{Nature
  communications}, vol.~10, no.~1, pp. 1--8, 2019.

\bibitem{bing2021toward}
Z.~Bing, A.~E. Sewisy, G.~Zhuang, F.~Walter, F.~O. Morin, K.~Huang, and
  A.~Knoll, ``Toward cognitive navigation: Design and implementation of a
  biologically inspired head direction cell network,'' \emph{IEEE Transactions
  on Neural Networks and Learning Systems}, 2021.

\bibitem{byrne2017learning}
J.~H. Byrne, \emph{Learning and memory: a comprehensive reference}.\hskip 1em
  plus 0.5em minus 0.4em\relax Academic Press, 2017.

\bibitem{bing2018end}
Z.~Bing, C.~Meschede, K.~Huang, G.~Chen, F.~Rohrbein, M.~Akl, and A.~Knoll,
  ``End to end learning of spiking neural network based on r-stdp for a lane
  keeping vehicle,'' in \emph{2018 IEEE international conference on robotics
  and automation (ICRA)}.\hskip 1em plus 0.5em minus 0.4em\relax IEEE, 2018,
  pp. 4725--4732.

\bibitem{lechner2020neural}
M.~Lechner, R.~Hasani, A.~Amini, T.~A. Henzinger, D.~Rus, and R.~Grosu,
  ``Neural circuit policies enabling auditable autonomy,'' \emph{Nature Machine
  Intelligence}, vol.~2, no.~10, pp. 642--652, 2020.

\bibitem{bing2020indirect}
Z.~Bing, C.~Meschede, G.~Chen, A.~Knoll, and K.~Huang, ``Indirect and direct
  training of spiking neural networks for end-to-end control of a lane-keeping
  vehicle,'' \emph{Neural Networks}, vol. 121, pp. 21--36, 2020.

\bibitem{milford2004ratslam}
M.~J. Milford, G.~F. Wyeth, and D.~Prasser, ``Ratslam: a hippocampal model for
  simultaneous localization and mapping,'' in \emph{IEEE International
  Conference on Robotics and Automation, 2004. Proceedings. ICRA'04. 2004},
  vol.~1.\hskip 1em plus 0.5em minus 0.4em\relax IEEE, 2004, pp. 403--408.

\bibitem{milford2013brain}
M.~J. Milford and A.~Jacobson, ``Brain-inspired sensor fusion for navigating
  robots,'' in \emph{2013 IEEE International Conference on Robotics and
  Automation}.\hskip 1em plus 0.5em minus 0.4em\relax IEEE, 2013, pp.
  2906--2913.

\bibitem{zhou2017brain}
S.-C. Zhou, R.~Yan, J.-X. Li, Y.-K. Chen, and H.~Tang, ``A brain-inspired slam
  system based on orb features,'' \emph{International Journal of Automation and
  Computing}, vol.~14, no.~5, pp. 564--575, 2017.

\bibitem{yu2019neuroslam}
F.~Yu, J.~Shang, Y.~Hu, and M.~Milford, ``Neuroslam: a brain-inspired slam
  system for 3d environments,'' \emph{Biological Cybernetics}, vol. 113, no.
  5-6, pp. 515--545, 2019.

\bibitem{milford2008mapping}
M.~J. Milford and G.~F. Wyeth, ``Mapping a suburb with a single camera using a
  biologically inspired slam system,'' \emph{IEEE Transactions on Robotics},
  vol.~24, no.~5, pp. 1038--1053, 2008.

\bibitem{tian2013rgb}
B.~Tian, V.~A. Shim, M.~Yuan, C.~Srinivasan, H.~Tang, and H.~Li, ``Rgb-d based
  cognitive map building and navigation,'' in \emph{2013 IEEE/RSJ International
  Conference on Intelligent Robots and Systems}.\hskip 1em plus 0.5em minus
  0.4em\relax IEEE, 2013, pp. 1562--1567.

\bibitem{davies2018loihi}
M.~Davies, N.~Srinivasa, T.-H. Lin, G.~Chinya, Y.~Cao, S.~H. Choday, G.~Dimou,
  P.~Joshi, N.~Imam, S.~Jain \emph{et~al.}, ``Loihi: A neuromorphic manycore
  processor with on-chip learning,'' \emph{Ieee Micro}, vol.~38, no.~1, pp.
  82--99, 2018.

\bibitem{furber2014spinnaker}
S.~B. Furber, F.~Galluppi, S.~Temple, and L.~A. Plana, ``The spinnaker
  project,'' \emph{Proceedings of the IEEE}, vol. 102, no.~5, pp. 652--665,
  2014.

\bibitem{kreiser2018pose}
R.~Kreiser, A.~Renner, Y.~Sandamirskaya, and P.~Pienroj, ``Pose estimation and
  map formation with spiking neural networks: towards neuromorphic slam,'' in
  \emph{2018 IEEE/RSJ International Conference on Intelligent Robots and
  Systems (IROS)}.\hskip 1em plus 0.5em minus 0.4em\relax IEEE, 2018, pp.
  2159--2166.

\bibitem{kreiser2018neuromorphic}
R.~Kreiser, M.~Cartiglia, J.~N. Martel, J.~Conradt, and Y.~Sandamirskaya, ``A
  neuromorphic approach to path integration: a head-direction spiking neural
  network with vision-driven reset,'' in \emph{2018 IEEE international
  symposium on circuits and systems (ISCAS)}.\hskip 1em plus 0.5em minus
  0.4em\relax IEEE, 2018, pp. 1--5.

\bibitem{grisetti2007improved}
G.~Grisetti, C.~Stachniss, and W.~Burgard, ``Improved techniques for grid
  mapping with rao-blackwellized particle filters,'' \emph{IEEE transactions on
  Robotics}, vol.~23, no.~1, pp. 34--46, 2007.

\bibitem{steckel2013batslam}
J.~Steckel and H.~Peremans, ``Batslam: Simultaneous localization and mapping
  using biomimetic sonar,'' \emph{PloS one}, vol.~8, no.~1, p. e54076, 2013.

\bibitem{elfes1989using}
A.~Elfes, ``Using occupancy grids for mobile robot perception and navigation,''
  \emph{Computer}, vol.~22, no.~6, pp. 46--57, 1989.

\bibitem{barry2006boundary}
C.~Barry, C.~Lever, R.~Hayman, T.~Hartley, S.~Burton, J.~O'Keefe, K.~Jeffery,
  and N.~Burgess, ``The boundary vector cell model of place cell firing and
  spatial memory,'' \emph{Reviews in the Neurosciences}, vol.~17, no. 1-2, pp.
  71--98, 2006.

\bibitem{samsonovich1997path}
A.~Samsonovich and B.~L. McNaughton, ``Path integration and cognitive mapping
  in a continuous attractor neural network model,'' \emph{Journal of
  Neuroscience}, vol.~17, no.~15, pp. 5900--5920, 1997.

\bibitem{ball2013openratslam}
D.~Ball, S.~Heath, J.~Wiles, G.~Wyeth, P.~Corke, and M.~Milford, ``Openratslam:
  an open source brain-based slam system,'' \emph{Autonomous Robots}, vol.~34,
  no.~3, pp. 149--176, 2013.

\bibitem{quigley2009ros}
M.~Quigley, K.~Conley, B.~Gerkey, J.~Faust, T.~Foote, J.~Leibs, R.~Wheeler, and
  A.~Y. Ng, ``Ros: an open-source robot operating system,'' in \emph{ICRA
  workshop on open source software}, vol.~3, no. 3.2.\hskip 1em plus 0.5em
  minus 0.4em\relax Kobe, 2009, p.~5.

\bibitem{kummerle2009measuring}
R.~K{\"u}mmerle, B.~Steder, C.~Dornhege, M.~Ruhnke, G.~Grisetti, C.~Stachniss,
  and A.~Kleiner, ``On measuring the accuracy of slam algorithms,''
  \emph{Autonomous Robots}, vol.~27, no.~4, pp. 387--407, 2009.

\end{thebibliography}

\end{document}